\documentclass[letterpaper, 10 pt, conference]{ieeeconf}

\IEEEoverridecommandlockouts

\UseRawInputEncoding

\usepackage{amsmath,amsfonts}
\usepackage{algorithmic}
\usepackage{algorithm}
\usepackage{array}
\usepackage[caption=false,font=normalsize,labelfont=sf,textfont=sf]{subfig}
\usepackage{textcomp}
\usepackage{stfloats}
\usepackage{url}
\usepackage{verbatim}
\usepackage{graphicx}
\usepackage{cite}
\usepackage{hyperref}
\usepackage{booktabs}
\usepackage{multirow}
\usepackage{doi}
\usepackage{threeparttable}
\usepackage{bbding}
\usepackage{amsmath}
\usepackage{float}
\usepackage{mathrsfs}
\usepackage{color}
\usepackage{comment}
\usepackage{makecell}
\usepackage{bm}
\usepackage{balance}

\title{\LARGE \bf
SWCF-Net: Similarity-weighted Convolution and Local-global Fusion for Efficient Large-scale Point Cloud Semantic Segmentation
}

\author{Zhenchao Lin$^{1}$, Li He$^{2}\textsuperscript{*}$, Hongqiang Yang$^{3}$, Xiaoqun Sun$^{3}$, Guojin Zhang$^{3}$, \\Weinan Chen$^{1}$, Yisheng Guan$^{1}$ and Hong Zhang$^{2}$, \IEEEmembership{Fellow, IEEE} % Albert Author$^{1}$ and Bernard D. Researcher$^{2}$% <-this % stops a space
\thanks{\textsuperscript{*}Corresponding author (hel@sustech.edu.cn)}
\thanks{$^{1}$Zhenchao Lin, Weinan Chen, Yisheng Guan are with the Biomimetic and Intelligent Robotics Lab (BIRL), School of Electromechanical Engineering, Guangdong University of Technology, Guangzhou, China.}%
\thanks{$^{2}$Li He and Hong Zhang ({hel, hzhang}@sustech.edu.cn) are with the Department of Electronic and Electrical Engineering, Southern University of Science and Technology, China, and with Shenzhen Key Laboratory of Robotics and Computer Vision.}
\thanks{$^{3}$Hongqiang Yang, Xiaoqun Sun and Guojin Zhang are with Meituan Technology Co., Ltd, Shenzhen, China.}
}

\begin{document}

\maketitle
\thispagestyle{empty}
\pagestyle{empty}

\begin{abstract}

Large-scale point cloud consists of a multitude of individual objects, thereby encompassing rich structural and underlying semantic contextual information, resulting in a challenging problem in efficiently segmenting a point cloud. Most existing researches mainly focus on capturing intricate local features without giving due consideration to global ones, thus failing to leverage semantic context. In this paper, we propose a Similarity-Weighted Convolution and local-global Fusion Network, named SWCF-Net, which takes into account both local and global features. We propose a Similarity-Weighted Convolution (SWConv) to effectively extract local features, where similarity weights are incorporated into the convolution operation to enhance the generalization capabilities. Then, we employ a downsampling operation on the $\bm{\mathit{K}}$ and $\bm{\mathit{V}}$ channels within the attention module, thereby reducing the quadratic complexity to linear, enabling Transformer to deal with large-scale point cloud. At last, orthogonal components are extracted in the global features and then aggregated with local features, thereby eliminating redundant information between local and global features and consequently promoting efficiency. We evaluate SWCF-Net on large-scale outdoor datasets SemanticKITTI and Toronto3D. Our experimental results demonstrate the effectiveness of the proposed network. Our method achieves a competitive result with less computational cost, and is able to handle large-scale point clouds efficiently. The code is available at https://github.com/Sylva-Lin/SWCF-Net

\end{abstract}

\section{INTRODUCTION}

Efficient large-scale point cloud semantic segmentation has garnered significant attention due to its remarkable performance in applications such as autonomous driving, augmented reality, and 3D scene understanding. Despite the rich spatial information provided by 3D point clouds, the irregular and unordered points present challenges for semantic segmentation. In recent years, deep learning approaches\cite{guo2020deep} have been employed for point cloud semantic segmentation, including point-based approaches such as PointNet++\cite{qi2017pointnet++} and PointMLP\cite{ma2022rethinking}. Such methods exhibit promising performance on small-scale point clouds. However, they mostly focus on extracting local features and cannot achieve satisfactory segmentation accuracy when extended to large-scale point clouds.

\begin{figure}[t]
	\centering
	\includegraphics[width=1.0\linewidth]{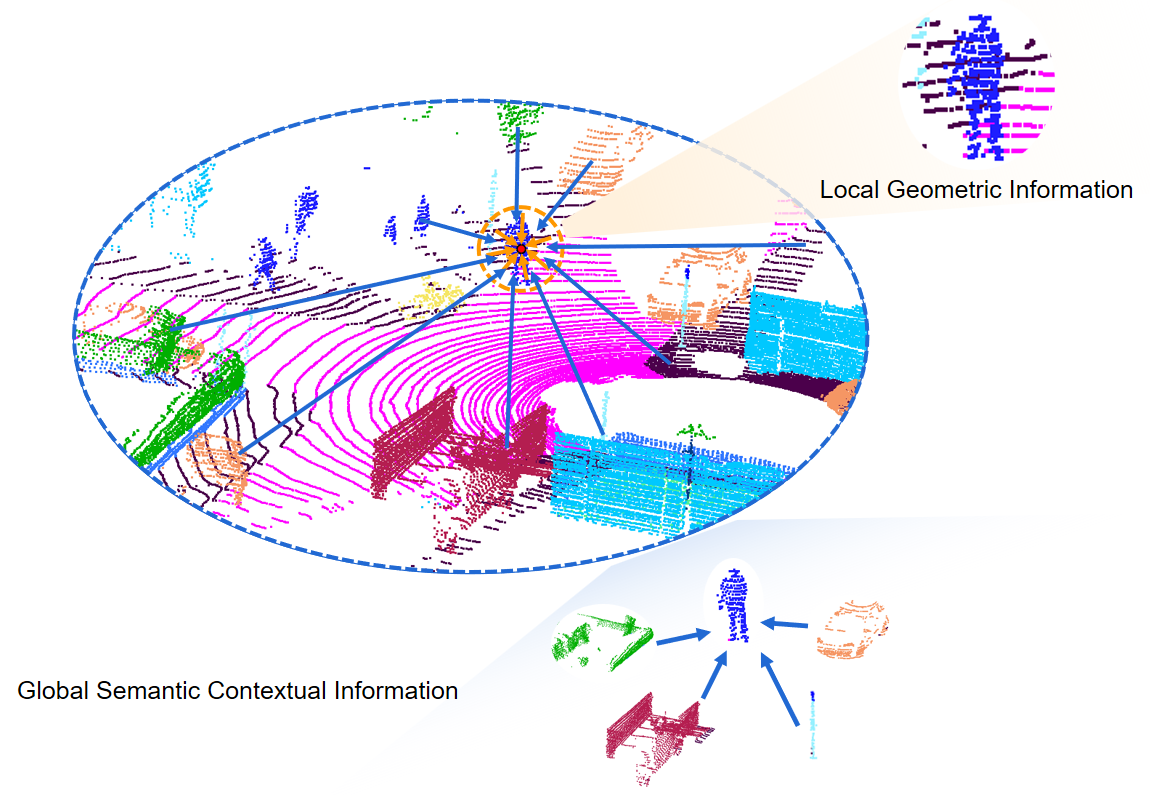}
	\caption{Demonstration of 3D LiDAR segmentation. In large-scale point clouds, semantic context is essential for segmentation. Many existing segmentation methods employ local features only and do not consider the global information, which is highly related with the semantic context. }
	\label{fig_1}
\end{figure}

Several recent works have begun to tackle the challenge of directly processing large-scale point clouds, such as SPG\cite{landrieu2018large}, RandLA-Net\cite{hu2021learning}, BAAF-Net\cite{qiu2021semantic}. Although these methods can handle large-scale point clouds, they still lack consideration of global features, resulting in suboptimal segmentation. As shown in Figure \textcolor[rgb]{1,0,0}{1}, for large-scale point cloud processing, local features can capture the geometric structure, while global features establish long-range dependency that harness underlying semantic context among objects, such as pedestrians typically found on sidewalks and cars generally on roads, etc. Several works\cite{ruan2023combining}, \cite{yan2020pointasnl} have also demonstrated the advantages of combining local and global features. However, current methods largely focus on extracting intricate local features while overlooking the utilization of global features.

In this paper, we propose a Similarity-Weighted Convolution and local-global Fusion Network, named SWCF-Net, which takes into account geometric and semantic context information. We propose SWConv for extracting local feature along with a low computational complexity Average Transformer designed to extract global features. Additionally, to integrate local and global features, we adopt an orthogonal fusion strategy. This involves extracting components from the global features orthogonal to the local ones and then aggregating both. This process eliminates redundant information that shares the same feature direction, thereby promoting a more effective learning process in the network.

In summary, the main contributions of this paper are as follows:

\begin{itemize}
\item We propose a Similarity-Weighted Convolution (SWConv) to achieve a low Gaussian complexity, which consequently improves the generalization ability of 3D convolutional operations and thereby yields improved local features.
\item We perform downsampling operations on the $K$ and $V$ channels of the attention module to accelerate the Transformer-based global encoder and fuse these global with local features through orthogonal fusion.
\item The proposed SWCF-Net has demonstrated competitive performance on both the SemanticKITTI and Toronto3D datasets while requiring lower computational resources to achieve higher segmentation accuracy than its counterparts.
\end{itemize}

\section{RELATED WORKS}

\subsection{Point Based Methods}

The point-based method can directly operate on point clouds. PointNet\cite{qi2017pointnet} utilizes several shared multi-layer perceptrons to extract pointwise features, which are then aggregated into global features using symmetric pooling functions. This allows for efficient point cloud semantic segmentation. However, due to its lack of consideration for local features, it does not achieve higher levels of segmentation precision. Subsequent to this, numerous works have proposed various local feature extractors.

PointNet++\cite{qi2017pointnet++} employs farthest point sampling (FPS) and nearest neighbor search (KNN) to group the points within a point cloud, thereby enabling the acquisition of local features. DGCNN\cite{wang2019dynamic} updates the features of query points by constructing a graph structure and leveraging the node and edge features within local graphs. KPConv\cite{thomas2019kpconv} proposes a deformable convolutional operator that computes the convolution kernel function based on the distances between kernel points and points in local regions of the point cloud, subsequently applying it to the corresponding local point cloud areas. Point Transformer\cite{zhao2021point} applies the attention mechanism to local regions and utilizes vector attention to compute weight matrices, thereby reducing computational costs. While these methods have demonstrated promising performance on small-scale point clouds, they are largely constrained by their downsampling techniques and computational approaches, which hinders their direct scalability to large-scale point clouds.

\subsection{Large-scale Point Clouds Methods}

To accelerate the processing speed of point clouds without compromising scene information, several methods have been designed for handling large-scale point clouds. SPG\cite{landrieu2018large} preprocesses large point clouds into superpoint graphs to facilitate learning the semantics of each superpoint. TangentCov\cite{tatarchenko2018tangent} proposes tangent convolutions by projecting local point clouds onto tangent planes and subsequently performing convolutional operations on the projected patterns. While these methods can handle large-scale point clouds, they require a complex preprocessing stage, which renders them inefficient for point cloud processing.

To enhance the segmentation efficiency. RandLA-Net\cite{hu2021learning} first adopts random sampling as the downsampling strategy among its encoding layers and introduces Local Spatial Encoding Modules along with Attention Pooling Modules to mitigate against losing valuable information caused by random sampling. BAAF-Net\cite{qiu2021semantic} introduces Bilateral Context Modules and Adaptive Fusion Modules to effectively encode and integrate features within local regions. DCNet\cite{yin2023dcnet} proposes Dual Attention Modules and Consistency Constraint Loss, which enhance the segmentation performance at point cloud boundaries. While these methods can efficiently process point clouds, their encoding modules only consider local feature information without capturing long-range dependencies among the points. Consequently, they can only deliver suboptimal segmentation performance. In large-scale point clouds, there exist underlying semantic context relationships between different object categories, and harnessing this information can aid the network in achieving better performance.

\subsection{Local and Global Fusion Methods}

In order to account for global long-range dependencies, some methods adopt a combined approach that integrates both local and global information for segmentation. PointASNL\cite{yan2020pointasnl} introduces a Local Non-Local (L-NL) module, where the local module utilizes traditional convolution operations, while the global module employs traditional attention mechanisms to compute global features. While this effectively harnesses the underlying information within point clouds, it must be noted that traditional attention mechanisms have quadratic complexity, which in turn necessitates substantial computational resources for this method. To capture long-range dependencies while minimizing computational costs, LGGCM\cite{du2022novel} employs the global attention mechanism only in its last few network layers. Additionally, it introduces a gated unit to effectively integrate feature information from both anterior and posterior layers. While the computational resources required at the bottom layers of the network are relatively lower due to the smaller number of point clouds, the feature information in these layer's point clouds is quite rich. Hence, executing global association in these layers does not lead to substantial improvements in accuracy. LG-Net\cite{zhao2023large} operates in a similar fashion to this method, with the distinction that it confines its global association process only to the last layer of its encoder. To address the issue of global computational complexity while fully exploiting long-range dependency capabilities, this paper introduces a lightweight attention module called Average Transformer. This module is employed in every encoding layer, enabling the network to achieve improved performance.

\begin{figure*}[t]
	\centering
	\includegraphics[width=\linewidth]{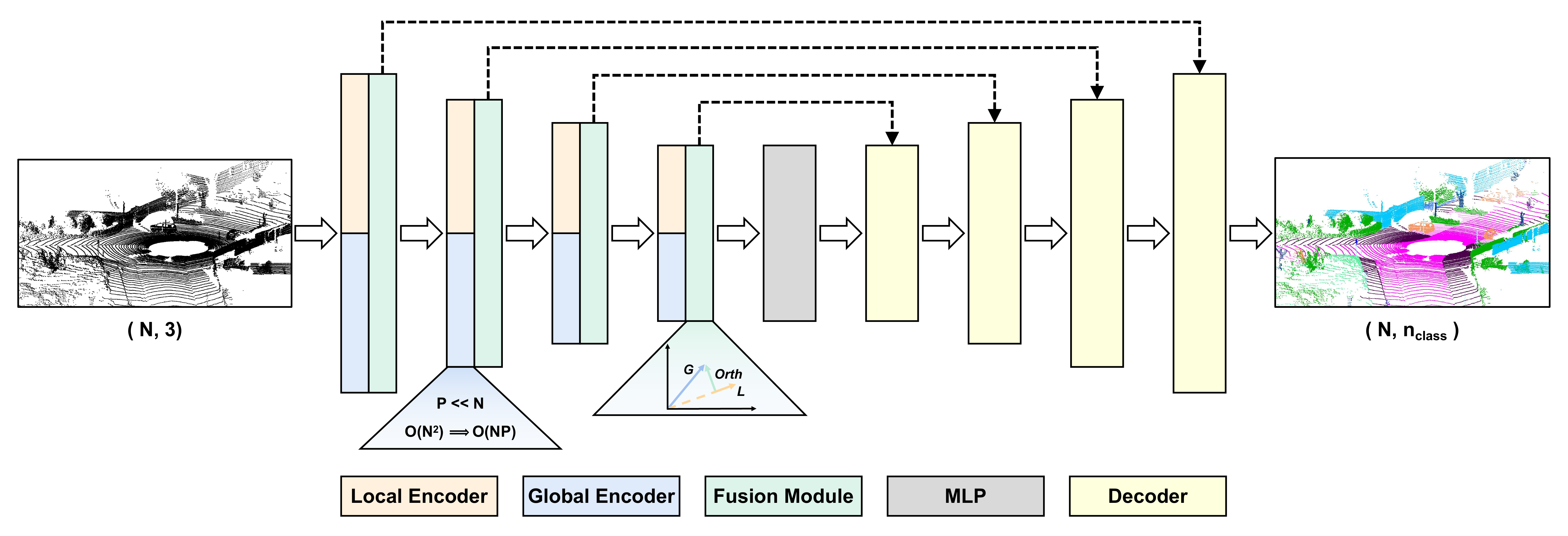}
	\caption{The architecture of our SWCF-Net. Our model adopts an encoder-decoder architecture, where the encoder consists of a local encoder, global encoder, and fusion module. The local encoder adopts the proposed similarity-weighted 3D CNN to capture local features. The global encoder uses a lightweight Transformer to capture global features. The fusion module employs orthogonal fusion to effectively integrate both local and global features.}
	\label{fig_2}
\end{figure*}

\section{METHODOLOGY}

Our SWCF-Net is constructed based on the encoder-decoder architecture, where the encoder component consists of three major parts: local encoder, global encoder, and fusion module, as shown in Figure \textcolor[rgb]{1,0,0}{2}.

\subsection{Local Encoder}

\begin{comment}
Convolutional neural networks (CNN) have achieved remarkable success in the field of computer vision, effectively enabling the extraction of local features from pixels\cite{kirillov2019panoptic}. The primary reason is that a smaller Gaussian complexity enables the attainment of better generalization capability\cite{bartlett2002rademacher}. The convolution operation can be approximated as an empirical Gaussian complexity, and pixels within a local region often exhibit high color similarity, resulting in a smaller Gaussian complexity\cite{li2016filter}. Consequently, traditional convolution operations can achieve better generalization within local regions (e.g., 3x3, 5x5, and 7x7).
\end{comment}

CNN is widely used in image object detection and segmentation\cite{ren2015faster}, \cite{strudel2021segmenter}. However, since point clouds are discrete, traditional convolutions cannot directly process a point cloud. To overcome this issue, some works\cite{thomas2019kpconv}, \cite{wu2019pointconv} employ MLP to approximate a convolutional kernel function based on 3D positions,

\begin{equation}
	\begin{aligned}
		Conv(g, \psi) = \sum_{i=0}^{N}\sum_{j=0}^{K} g(x_j - x_i)\psi(f_i)
	\end{aligned}
\end{equation}

\noindent where $x$ and $f$ respectively represent the position $\mathcal{P} \in \mathbb{R}^{N \times 3}$ and feature  $\mathcal{F} \in \mathbb{R}^{N \times D}$ of $N$ points. $K$ represents the number of neighbors for a query point, and $g(\cdot)$ and $\psi(\cdot)$ are two scalar functions approximated by MLP. 

Despite its success in 2D image tasks, CNN always shows low generalization on 3D point cloud tasks. One critical problem is the low similarity of a local 3D region used in point cloud. The generalization capability of can be characterized using Gaussian or Rademacher complexities\cite{bartlett2002rademacher}. In \cite{li2016filter}, the empirical Gaussian complexity $\hat{G}_N(F)$ of the function class $F$ is defined as

\begin{equation}
	\begin{aligned}
		\hat{G}_N(F) \leq C \mathop{max}\limits_{j, j' \in \mathcal{N}}\sqrt{\sum_{i=1}^{N}\Vert x_i(j)-x_i(j')\Vert^2}
	\end{aligned}
\end{equation}

\noindent where $F$ represents the class computed by a CNN consisting of one convolutional layer and one fully connected layer, $C$ is a constant, $x_i(j)$ is the $j$-th element of vector $x_i$, and $\mathcal{N}$ is a local region. As a main result of\cite{bartlett2002rademacher}, a classifier learnt from function class $F$ with a low Gaussian complexity are likely to have a high generalization ability. Unlike 2D images, the local regions of point clouds often contain points from different categories, leading to a low local similarity and consequently high Gaussian complexity. Hence, general 3D convolutional operations often fail to work on wild cases dissimilar to the training set.

\begin{comment}
As shown in \cite{li2016filter}, CNN can be approximated as an empirical Gaussian model, whose complexity is related to the similarity within that local region. A higher similarity generally corresponds to a low complexity, indicating good generalization capabilities \cite{bartlett2002rademacher}. Unlike 2D images, the local regions of point clouds often contain points from different categories.
\end{comment}

The main idea to tackle this issue lies in the pursuit of a low Gaussian complexity, or as its empirical metric, a high local similarity. We propose the Similarity-Weighted Convolution (SWConv), as shown in Figure \textcolor[rgb]{1,0,0}{3}. SWConv attempts to eliminate points that are irrelevant to the query point. Rather than a hard threshold for binary partition, we adopt a weighted convolution framework to prevent a hard classifier. The similarity-weighted convolution then is 
\begin{equation}
	\begin{aligned}
		SWConv(g, \omega, \psi) = \sum_{i=0}^{N}\sum_{j=0}^{K} g(x_j - x_i)\omega(f_i, f_j)\psi(f_i) \\
		\omega(f_i, f_j) = Softmax(\varphi(f_j - f_i)) \ \ \ \ \ \ \ \ \ \ \ \ \ \
	\end{aligned}
\end{equation}

\noindent where $\omega$ denotes the filter weight function, $\varphi$ represents the MLP operation. The difference between feature vectors $f_i$ and $f_j$ is passed through an MLP operation and subsequently a Softmax function to obtain a weight matrix, which is then applied to the local region features. In order to avoid introducing excessive resource consumption and computational burden, we utilize the feature vector difference $f_j - f_i$ between a local point and its corresponding query point to obtain the weights and adopt residual connections. High weights are assigned to the points with high similarity values. As a result, the CNN outputs are dominated by points with high similarity values, and the corresponding Gaussian complexity then is assumed to be low. Given the SWConv module, the local features is

\begin{equation}
	\begin{aligned}
		f_{local} = \alpha(SWConv(x, \beta(f))) + \gamma(f) \\
	\end{aligned}
\end{equation}

\noindent where $\alpha(\cdot), \beta(\cdot), \gamma(\cdot)$ are three scalar functions approximated by MLP to construct residual connections.

\begin{figure}[t]
	\centering
	\includegraphics[width=3.3in]{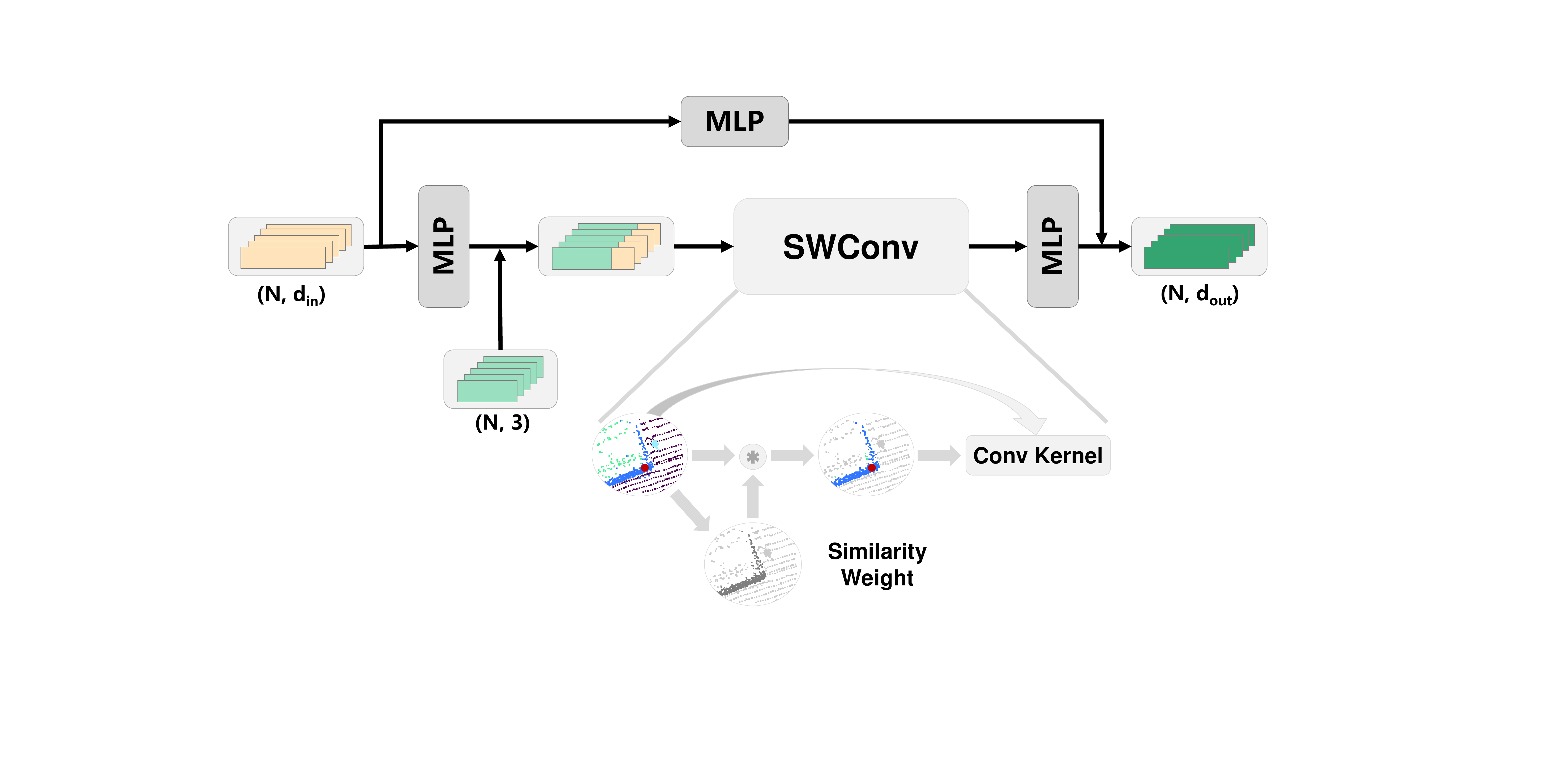}
	\caption{The proposed local encoding module. The Similarity-Weighted Convolution (SWConv) component is designed to apply weighted filtering to point cloud features at local regions.}
	\label{fig_3}
\end{figure}

\subsection{Global Encoder}

Many previous researches show that global features can improve object semantic segmentation by effectively capturing the semantic context among objects and establishing long-range dependencies. Several works\cite{ruan2023combining}, \cite{yan2020pointasnl} have also demonstrated the advantages of combining local and global features. As an efficient global feature extractor, Transformer\cite{vaswani2017attention} can effectively capture global long-range dependencies and is widely used in point cloud analysis. The attention mechanism of Transformer, however, exhibits quadratic complexity with respect to the number of input points, prohibiting its application on large-scale point clouds due to limited GPU memory and the demand of real-time processing. 

Linear Transformer such as low-rank estimation\cite{sharma2023truth} and rank recovery\cite{han2023flatten} are used in large language models and image processing for acceleration. However, the existing linear attention modules are not suitable for large-scale point clouds because point clouds are typically sparse and irregularly distributed. It is still a challenge to adopt Transformer in large-scale point cloud processing. Other than linear Transformer, downsampling is a popular in Transformer acceleration. For example, in  PVT2\cite{wang2022pvt}, the number of pixels $N$ for the $K$ and $V$ channels in attention is first downsampled through average pooling, leading to a reduction from $N$ to $P$.

\begin{figure}[t]
	\centering
	\includegraphics[width=1.0\linewidth]{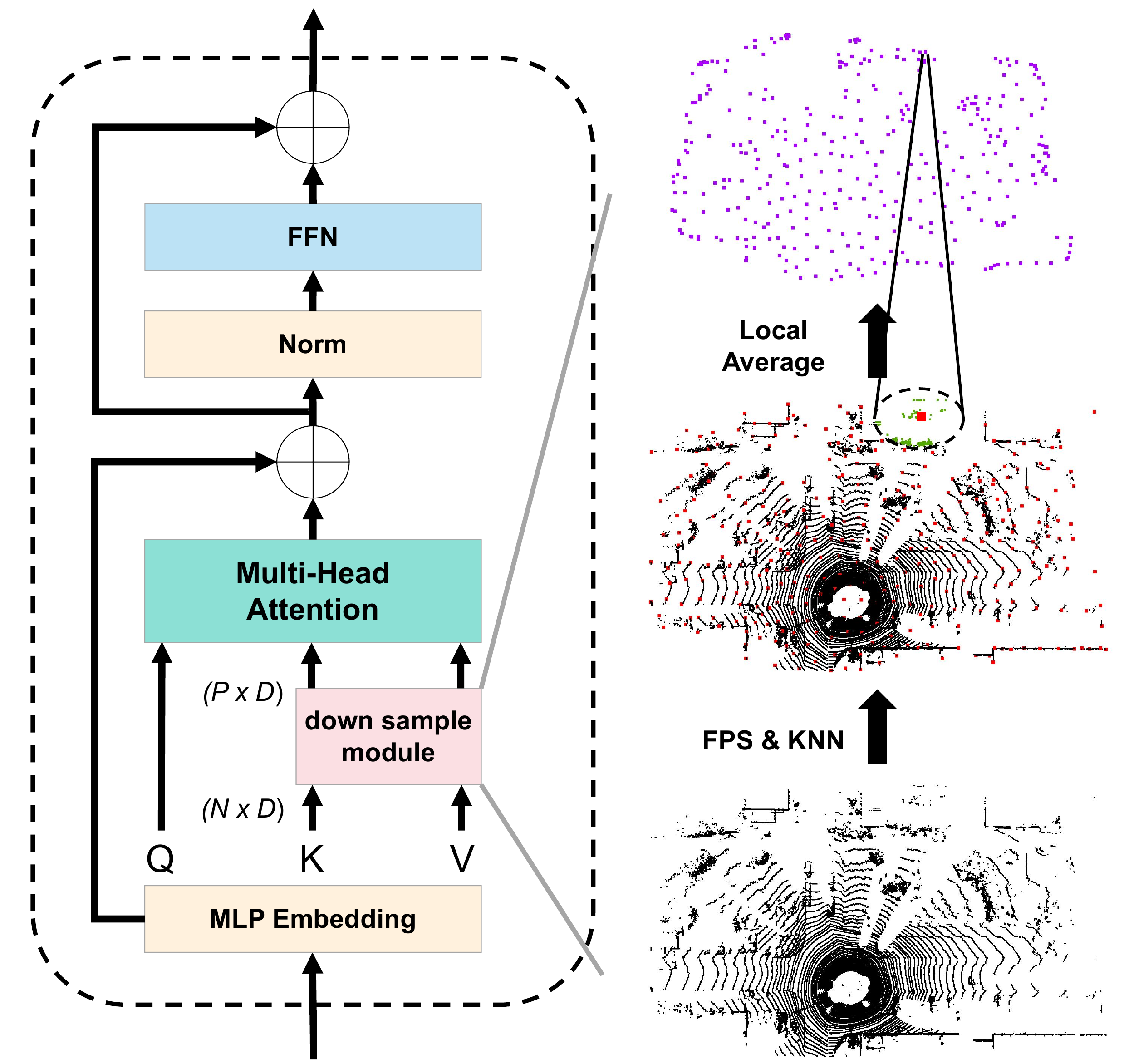}
	\caption{The proposed gloabl encoding module.  The Average Transformer relies on the down sample module to reduce the number of points in the $K$ and $V$ channels of the attention mechanism, from $N$ down to $P$, where $P \ll N$. Black points represent the raw point cloud, red points indicate the down-sampled points after FPS, green points denote neighboring points of the sampled points, and purple points represent the feature points obtained through averaging within the local area. Consequently, this enables the avoidance of the quadratic complexity inherent in traditional attention mechanisms.}
	\label{fig_4}
\end{figure}

Despite its success in 2D image downsampling, average pooling is not proper for point cloud. Point clouds lack a regular grid-like structure. Therefore, the number of in-voxel points after voxel downsampling, the analogue to its 2D counterpart, can not be guaranteed to be consistent, leading to the failure of the following attention operations. Although there exist several adaptive voxel sampling techniques with consistent output in point amount\cite{reinke2022locus}, the additional computational burden is always high. Other sampling methods, such as curvature-based sampling\cite{zhu2022curvature} and clustering-based sampling\cite{zhang2022patchformer}, are also shown to be time consuming. Random sampling, which is efficient, can not ensure a uniform distribution of sampled points.

As shown in Figure \textcolor[rgb]{1,0,0}{4}, our downsampling module begins by performing Farthest Point Sampling (FPS) on the input set of $N$ points. Subsequently, for each sampled point, we use KNN to find its $K$ nearest neighbors. Finally, the average feature of these $K$-neighbor features is computed, leading to a $P\times D$ sampled point feature matrix $A$ where $P$ is the desired number of points, which is a user-designed value to balance efficiency and accuracy, and $D$ is the dimension of features. We use the downsampled $P$ points in the $K$ and $V$ channels and push the original $N$ points to the $Q$ channel. Therefore, running Transformer on $P$ points, rather than the original $N$, can significantly reduce the computational complexity from $O(N^2)$ to $O(NP)$. Additionally, to capture information from various dimensions of the semantic space, we adopt a multi-head attention mechanism to learn global correlations,

\begin{equation}
\begin{aligned}
f_{global} = T + \mathcal{F}(T)  \ \ \ \ \ \ \ \ \ \ \ \ \ \ \ \ \ \ \ \ \ \ \ \ \ \ \ \ \\
T = X + \mathcal{MHA}(XW_Q, AW_K, AW_V)
\end{aligned}
\end{equation}

\noindent where $\mathcal{F}(\cdot)$ is a simple feed-forward layer, $\mathcal{MHA}(\cdot)$ represents the operation of the multi-head attention, $X \in \mathbb{R}^{N \times D}$ denotes features of the original input points $N$ , $A \in \mathbb{R}^{P \times D}$ represents the average features points within each local region. $W_Q, W_K$ and $W_V$ are linear projection parameters.

\subsection{Fusion Module}

In order to enable an effective fusion of the local features and the global features, we use orthogonal fusion, as shown in Figure \textcolor[rgb]{1,0,0}{5}. We perform an orthogonal projection of the global features onto the local features. To eliminate redundancy, we replace the global features by its components orthogonal to the local features. Finally, the local features are concatenated with the projected global components. The final output features are

\begin{equation}
\begin{aligned}
f_{orth} = f_{global} - \frac{f_{local}^{T}f_{global}}{\Vert f_{local} \Vert} \cdot f_{local} \\
f_{output} = \delta ([f_{local}, f_{orth}]) \ \ \ \ \ \ \ \ \ \ \ \ \ \ \
\end{aligned}
\end{equation}

\noindent where $\delta(\cdot)$ and $[\ ,\ ]$ respectively denote the MLP operation and concatenation.

\begin{table*}[t]
	\caption{Quantitative Results of Different Methods on SemanticKITTI}
	\label{table_example}
	\resizebox{\textwidth}{!}{
		\begin{tabular}{l|c|ccccccccccccccccccc}
			\toprule[1.5pt]
			Methods      & \rotatebox{90}{mIoU(\%)} & \rotatebox{90}{car} & \rotatebox{90}{bicycle} & \rotatebox{90}{motorcycle} & \rotatebox{90}{truck} & \rotatebox{90}{other-vehicle} & \rotatebox{90}{person} & \rotatebox{90}{bicyclist} & \rotatebox{90}{motorcylist} & \rotatebox{90}{road} & \rotatebox{90}{parking} & \rotatebox{90}{sidewalk} & \rotatebox{90}{other-ground} & \rotatebox{90}{building} & \rotatebox{90}{fence} & \rotatebox{90}{vegetation} & \rotatebox{90}{trunk} & \rotatebox{90}{terrain} & \rotatebox{90}{pole} & \rotatebox{90}{traffic-sign} \\ \midrule[0.5pt]
			PointNet\cite{qi2017pointnet}    & 14.6     & 46.3 & 1.3     & 0.3        & 0.1   & 0.8           & 0.2    & 0.2       & 0.0         & 61.6 & 15.8    & 35.7     & 1.4          & 41.4     & 12.9  & 31.0       & 4.6   & 17.6    & 2.4  & 3.7          \\
			SPG\cite{landrieu2018large}         & 17.4     & 49.3 & 0.2     & 0.2        & 0.1   & 0.8           & 0.3    & 2.7       & 0.1         & 45.0 & 0.6     & 28.5     & 0.6          & 64.3     & 20.8  & 48.9       & 27.2  & 24.6    & 15.9 & 0.8          \\
			PointNet++\cite{qi2017pointnet++}  & 20.1     & 53.7 & 1.9     & 0.2        & 0.9   & 0.2           & 0.9    & 1.0       & 0.0         & 72.0 & 18.7    & 41.8     & 5.6          & 62.3     & 16.9  & 46.5       & 13.8  & 30.0    & 6.0  & 8.9          \\
			TangentConv\cite{tatarchenko2018tangent} & 40.9     & 90.8 & 2.7     & 16.5       & 15.2  & 12.1          & 23.0   & 28.4      & 8.1         & 83.9 & 33.4    & 63.9     & 15.4         & 83.4     & 49.0  & 79.5       & 49.3  & 58.1    & 35.8 & 28.5         \\
			PointASNL\cite{yan2020pointasnl}   & 46.8     & 87.9 & 0.0     & 25.1       & 39.0  & 29.2          & 34.2   & 57.6      & 0.0         & 87.4 & 24.3    & 74.3     & 1.8          & 83.1     & 43.9  & 84.1       & 52.2  & \textbf{70.6}    & \textbf{57.8} & 36.9         \\
			RangeNet53++\cite{milioto2019rangenet++}      & 52.2     & 91.4 & 25.7    & 34.4       & 25.7  & 23.0          & 38.3   & 38.8      & 4.8        & \textbf{91.8} & \textbf{65.0}    & 75.2     & 27.8         & 87.4     & 58.6  & 80.5       & 55.1  & 64.6    & 47.9 & 55.9         \\
			SqueezeSegV3\cite{xu2020squeezesegv3}      & 55.9     & 92.5 & 38.7    & 36.5       & 29.6  & 33.0          & 45.6   & 46.2      & 20.1        & 91.7 & 63.4    & 74.8     & 26.4         & 89.0     & 59.4  & 82.0       & 58.7  & 65.4    & 49.6 & 58.9         \\
			RandLA-Net\cite{hu2021learning}  & 55.9     & 94.2 & 29.8    & 32.2       & \textbf{43.9}  & \textbf{39.1}          & 48.4   & 47.4      & 9.4         & 90.5 & 61.8    & 74.0     & 24.5         & 89.7     & 60.4  & 83.8       & 63.6  & 68.6    & 51.0 & 50.7         \\
			LG-Net\cite{zhao2023large}  & 56.3     & 93.9 & 36.5    & 36.0       & 42.7  & 36.8          & 50.1   & 47.7      & 0.9         & 90.6 & 63.0    & 74.2     & 19.1         & \textbf{90.6}     & 62.3 & 84.6       & 64.1  & 69.7    & 51.5 & 55.0         \\
			KPConv\cite{thomas2019kpconv}      & 58.8     & \textbf{95.0} & 30.2    & 42.5       & 33.4  & 44.3          & \textbf{61.5}   & \textbf{61.6}      & 11.8        & 90.3 & 61.3    & 72.7     & \textbf{31.5}         & 90.5     & \textbf{64.2}  & \textbf{84.8}       & \textbf{69.2}  & 69.1    & 56.4 & 47.4         \\ \midrule[0.5pt]
			SWCF-Net (Ours)        & \textbf{60.5}        & 94.6    & \textbf{51.8}       & \textbf{46.9}          & 34.3    & 37.3             & 59.4      & 61.0         & \textbf{29.9}           & 91.6    & 58.5       & \textbf{75.3}        & 31.2            & 89.2        & 60.5     & 82.2          & 62.7     & 67.8       & 54.0    & \textbf{61.6}            \\ \bottomrule[1.5pt]
		\end{tabular}
	}
\end{table*}

\section{EXPERIMENTS}

We report detailed experimental settings and evaluation results on two large-scale outdoor datasets, SemanticKITTI and Toronto3D. Subsequently, we conduct an extensive ablation studies to examine the effectiveness of individual components.

\begin{figure}[t]
	\centering
	\includegraphics[width=1.0\linewidth]{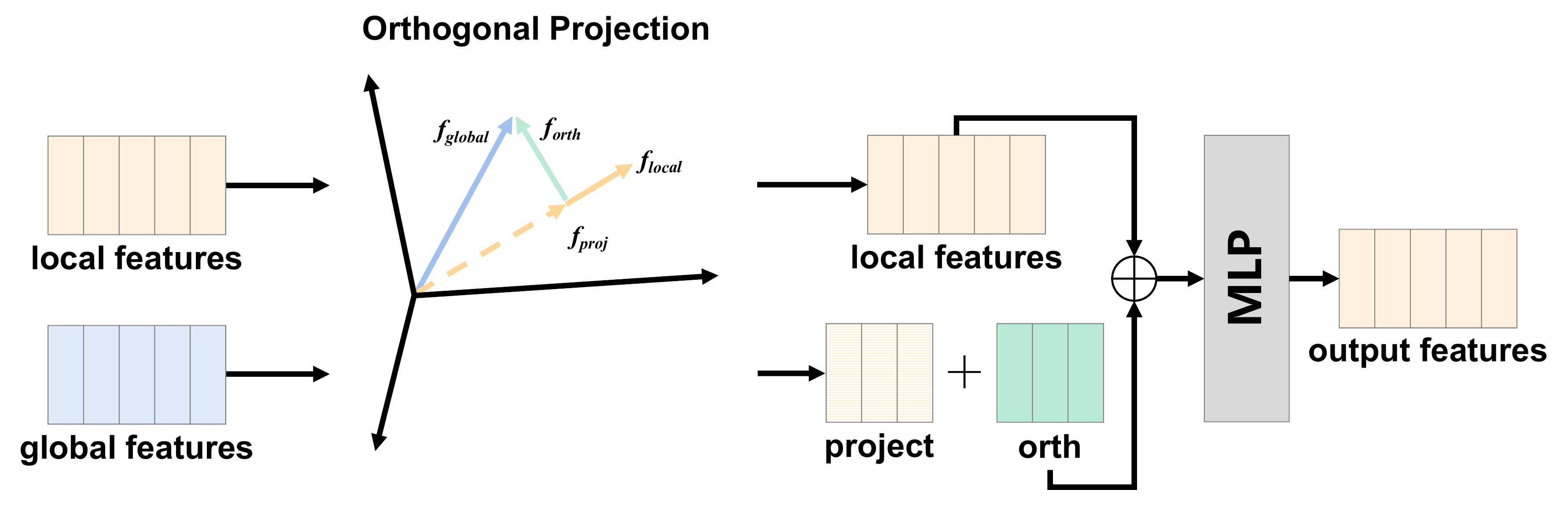}
	\caption{The proposed fusion module. Orthogonal components are extracted in global features and then aggregated with local features, thereby eliminating redundant information between local and global features.}
	\label{fig_5}
\end{figure}

\subsection{Dataset and Metric}

\textbf{SemanticKITTI.} SemanticKITTI\cite{behley2019semantickitti} is built upon the well-known KITTI Vision benchmark and designed to portray complex outdoor traffic scenarios. The dataset contains 22 stereo point cloud sequences, accumulating to a total of 43,552 LiDAR scans that are densely annotated. Each individual scan represents a large-scale point cloud consisting of approximately one hundred thousand points, spanning an area of about $160 \times 160 \times 20$ meters in real-world. The raw 3D points have been accurately labeled into 19 semantic classes. Officially, sequences ranging from 00 to 07 and from 09 to 10 (19,130 scans) are utilized for training, while sequence 08 (4071 scans) serves as the validation set. Lastly, sequences from 11 through 21 (20,351 scans) are designated for online testing.

\textbf{Toronto-3D.} Toronto-3D\cite{tan2020toronto} was collected using a 32-line LiDAR sensor in large-scale urban outdoor environments. It comprises over 78 million points, covering approximately 1 kilometer of road area. The dataset encompasses the following eight categories: Road, Road marking, Natural, Building, Utility line, Pole, Car, and Fence. For a fair comparison, we divided the dataset into four subsets: L001, L002, L003, L004, where L002 was used for testing.

\textbf{Metric.} We utilize the mean Intersection over Union (mIoU) as the evaluation metric to assess the performance. mIoU is calculated as 

\begin{equation}
\begin{aligned}
mIoU = \frac{1}{C} \sum_{c=1}^{C}(\frac{TP_c}{TP_c + FP_c + FN_c})
\end{aligned}
\end{equation}

\noindent where TP, FP, FN represent the number of True Positive, False Positive, and False Negative of classified points, respectively. $C$ refers to the number of classes in the dataset.

\subsection{Implementation Details}

We use the Adam optimizer to optimize our model, with an initial learning rate $10^{-2}$. When training on SemanticKITTI and Toronto-3D, the batch size was respectively set to 5 and 4. The network is trained for 100 epochs, with the learning rate decreasing by $5\%$ after each epoch. During the training stage, we sample a fixed number of $N=45,056$ points from the raw point cloud as input to the network. During the testing stage, the whole raw point cloud is fed into our network for inferring semantic labels without any pre-/post-processing. In order to address the issue of class imbalance, we employ a weighted cross-entropy as our loss function. For the Average Transformer, we set $P=176$ across all experiments. All experiments were conducted on a single RTX 3090.

\begin{table*}[t]
	\caption{Quantitative Results of Different Methods on Toronto3D}
	\label{table_example}
	\resizebox{\textwidth}{!}{
		\begin{tabular}{l|c|cccccccc}
			\toprule[1.0pt]
			Methods         & mIoU(\%) & Road  & Rd mrk. & Natural & Building & Util.line & Pole  & Car   & Fence \\ \midrule[0.3pt]
			PointNet++ \cite{qi2017pointnet++}     & 59.47    & 92.90 & 0.00    & 86.13   & 82.15    & 60.96     & 62.81 & 76.41 & 14.43 \\
			DGCNN \cite{wang2019dynamic}          & 61.79    & 93.88 & 0.00    & 91.25   & 80.39    & 62.40     & 62.32 & 88.26 & 15.81 \\
			MS-PCNN \cite{ma2019multi}  & 65.89    & 93.84 & 3.83    & 93.46   & 82.59    & 67.80     & 71.95 & 91.12 & 22.50 \\
			KPConv  \cite{thomas2019kpconv}        & 69.11    & 94.62 & 0.06    & 96.07   & 91.51    & 87.68     & 81.56 & 85.66 & 15.72 \\
			RandLA-Net \cite{hu2021learning}     & 77.71    & 94.61 & \textbf{42.62}   & 96.89   & 93.01    & 86.51     & 78.07 & 92.85 & 37.12 \\
			FGC-AFNet \cite{chen2023feature}     & 75.78    & 92.09 & 22.32   & 96.25   & 93.55    & 85.36     & 78.48 & 82.61 & 55.55 \\
			SWCF-Net (Ours)           & \textbf{78.51}        & \textbf{95.86}     & 17.93       & \textbf{98.14}       & \textbf{94.54}        & \textbf{87.70}         & \textbf{83.73}     & \textbf{94.08}     & \textbf{56.48}     \\ \midrule[0.3pt]
			RandLA-Net (RGB) \cite{hu2021learning} & 81.77    & 96.69 & 64.21   & 96.92   & 94.24    & 88.06     & 77.84 & 93.37 & 42.86 \\
			BAAF-Net (RGB) \cite{qiu2021semantic}   & 81.23    & \textbf{96.81} & \textbf{67.34}   & 96.83   & 92.20    & 86.87     & \textbf{82.31} & 93.14 & 34.02 \\
			FGC-AFNet (RGB) \cite{chen2023feature}  & 81.92    & 95.66 & 61.09   & 97.29   & \textbf{94.32}    & 86.89     & 81.74 & 88.27 & 50.08 \\
			SWCF-Net (Ours) (RGB)       & \textbf{82.58}    & 96.27 & 59.70   & \textbf{97.36}   & 94.30    & \textbf{88.31}     & 80.48 & \textbf{94.02} & \textbf{50.22} \\ \bottomrule[1.0pt]
		\end{tabular}
	}
\end{table*}

\subsection{Semantic Segmentation Results}

\begin{figure}[t]
	\centering
	\includegraphics[width=1.0\linewidth]{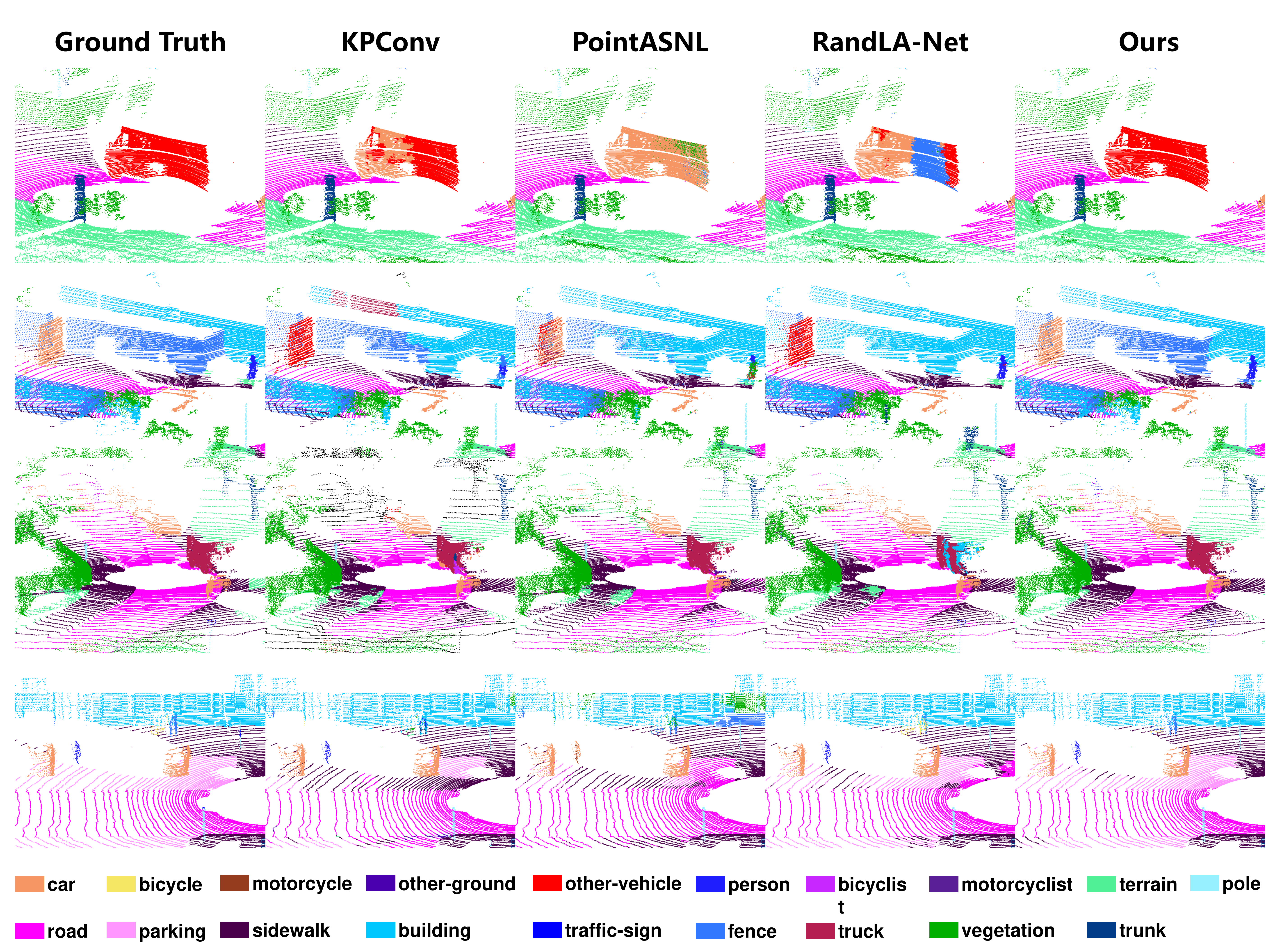}
	\caption{Qualitative results of different approaches on the validation set of SemanticKITTI}
	\label{fig_6}
\end{figure}

\textbf{SemanticKITTI.} We compared the results of SWCF-Net and existing methods on the test set of SemanticKITTI, as shown in Table \textcolor[rgb]{1,0,0}{\uppercase\expandafter{\romannumeral1}}. SWCF-Net demonstrates competitive performance against previous methods and surpasses them in several categories. Our SWCF-Net outperforms the baseline method (RandLA-Net) by a large margin of $4.6\%$. The significant improvements in accuracy are primarily observed in small object categories such as bicycle, motorcycle, bicyclist, and motorcyclist. This is attributed to our proposed method where the Convolution-Transformer fusion effectively leverages both the underlying geometric structure and semantic context information in large-scale point clouds. This allows the network to learn these challenging small categories, thereby resulting in an improvement of overall accuracy. 

We have compared the segmentation effects of SWCF-Net with several related methods, as shown in Figure \textcolor[rgb]{1,0,0}{6}. In the first two rows of Figure \textcolor[rgb]{1,0,0}{6}, other methods confuse vehicles with cars, while our approach can accurately segment them. This is mainly attributed to our Local Encoding Module, which effectively extracts the local geometric features of point clouds. In the last two rows of Figure \textcolor[rgb]{1,0,0}{6}, our method distinguishs between different ground categories such as road, parking, and sidewalk while lots errors are found in competing methods. This improved performance is facilitated by the introduction of global features, which enhance the segmentation effect for larger-scale objects. 

To compare the overall scene segmentation performance, we have conducted a comparison against the Baseline across entire scenes, as shown in Figure \textcolor[rgb]{1,0,0}{7}. The red highlighted areas represent errors in the segmentation task, both our method and the Baseline approach exhibit errors in the same regions. However, our method significantly reduces the number of incorrect labels. The key lies in our incorporation of global features, which enables the network to leverage underlying semantic context information for enhanced learning.

\textbf{Toronto-3D.} We present the semantic segmentation results on the Toronto-3D dataset, as shown in Table \textcolor[rgb]{1,0,0}{\uppercase\expandafter{\romannumeral2}}. This dataset incorporates RGB information, hence we conducted two separate sets of experiments to explore the effects of with/without RGB data. Compared to several recent state-of-the-art approaches, our approach demonstrates a more competitive performance in terms of results. For results without RGB, our method achieves the highest IoU in six out of eight categories, and once RGB is available, our obtains the highest IoU in four categories, while achieving the overall highest mIoU in both cases. This indicates that the proposed method exhibits robustness in large urban scenes, effectively capturing both the local geometric structures and long-range dependencies within point clouds.

\begin{figure}[t]
	\centering
	\includegraphics[width=1.0\linewidth]{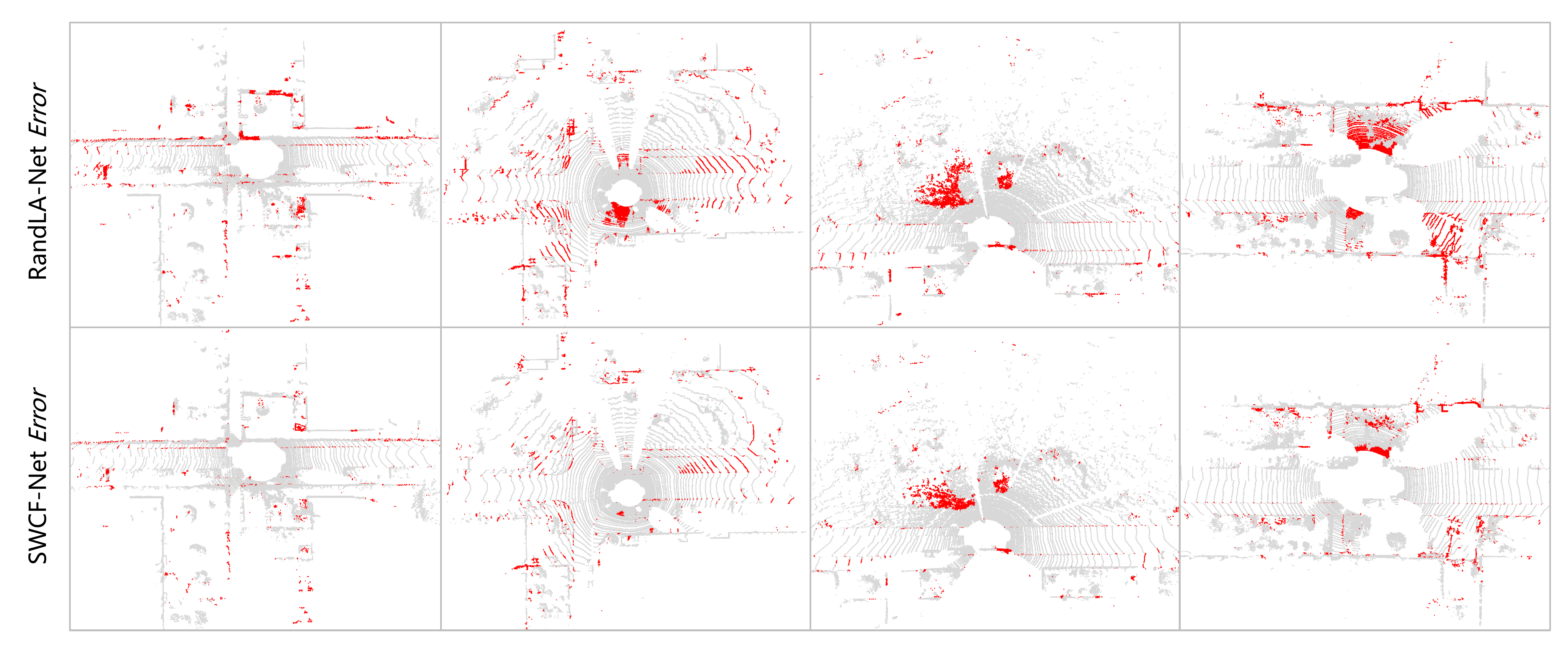}
	\caption{A demonstration of segmentation errors. Top: Red points denote segmentation errors for RandLA-Net. Bottom: Red points denote segmentation errors for our SWCF-Net.}
	\label{fig_7}
\end{figure}

\subsection{Ablation Studies}

In this subsection, we conduct ablation studies to validate the effectiveness of each proposed module. All ablation studies are performed on the validation set of SemanticKITTI and employ mIoU as the evaluation metric.

\textbf{Effectiveness of different modules.} We illustrate the effectiveness of different modules in Table \textcolor[rgb]{1,0,0}{\uppercase\expandafter{\romannumeral3}}. We first conduct an ablation on the local encoding module. In RandLA-Net\cite{hu2021learning}, two LocSE modules and an attentive pooling are stacked as the local encoding module. We replace with our proposed SWConv introduced in Section \uppercase\expandafter{\romannumeral3}-A. As can be seen from Table \textcolor[rgb]{1,0,0}{\uppercase\expandafter{\romannumeral3}}, the mIoU shows a $1.9\%$ increase. Next, we build upon this by incorporating our Average Transformer proposed in Section \uppercase\expandafter{\romannumeral3}-B, and fuse the two types of features using a simple concatenation. The results demonstrate that the integration of global features yields an improvement by $1.8\%$. Finally, to facilitate a more effective integration of local and global features, we employ the Orthogonal Fusion module proposed in Section \uppercase\expandafter{\romannumeral3}-C. Compared to simple concatenation, this module effectively eliminates redundant information between the two types of features, contributing to improved performance of the network.

%\begin{table}[t]
%	\caption{Ablation Studies for Different Modules}
%	\label{table_example}
%	\resizebox{1.0\linewidth}{!}{
%	\begin{tabular}{l|cccc|c}
%		\toprule[1.0pt]
%		Architecture & \rotatebox{90}{BFConv} & \rotatebox{90}{Centroid Transformer} & \rotatebox{90}{Concatenation} & \rotatebox{90}{Orthogonal Fusion} & mIoU \\ \midrule[0.3pt]
%		Baseline (RandLA-Net\cite{hu2021learning}) & \XSolidBrush & \XSolidBrush & \XSolidBrush & \XSolidBrush & 57.1 \\ \midrule[0.3pt]
%		\multirow{3}{*}{proposed modules} & \Checkmark & \XSolidBrush & \XSolidBrush & \XSolidBrush & 59.0 \\
%		 & \Checkmark & \Checkmark & \Checkmark & \XSolidBrush & 60.8 \\
%		 & \Checkmark & \Checkmark & \XSolidBrush & \Checkmark & 61.6 \\ \bottomrule[1.0pt]
%	\end{tabular}
%	}
%\end{table}

\begin{table}[t]
	\caption{Ablation Studies for Different Modules}
	\label{table_example}
	\resizebox{1.0\linewidth}{!}{
		\begin{tabular}{c|c}
			\toprule[1.0pt]
			Model & mIoU (\%) \\ \midrule[0.3pt]
			Baseline (RandLA-Net\cite{hu2021learning}) & 57.1 \\ \rule{0pt}{10pt}
			replace local feature aggregation module with SWConv & 59.0 \\ \rule{0pt}{10pt}
			add Average Transformer with Concatenation & 60.8 \\ \rule{0pt}{10pt}
			add Average Transformer with Orthogonal Fusion & 61.6 \\ \bottomrule[1.0pt]
		\end{tabular}
	}
\end{table}

\textbf{Effectiveness of global encoder in different layers.} To demonstrate the effectiveness of the global encoder at each layer within the encoding stage, we conducted ablation studies as shown in Table \textcolor[rgb]{1,0,0}{\uppercase\expandafter{\romannumeral4}}. We obtain higher mIoU when we employ more layers in the global encoding. Due to limitations imposed by the computationally expensive global encoder in previous works, global information was only utilized at the last layer of the network, which failed to adequately capture long-range dependencies. Our proposed Average Transformer possesses a lightweight solution.

%\begin{table}[t]
%	\caption{Ablation Studies on the Global Encoder in Different Encoding Layers}
%	\label{table_example}
%	\resizebox{1.0\linewidth}{!}{
%		\begin{tabular}{ccccc}
%			\toprule[1.0pt]
%			\multicolumn{4}{c}{Encoding layer} & \ \\ \cline{1-4} \rule{0pt}{10pt}
%			First layer & Second layer & Third layer & Fourth layer & mIoU \\ \hline \rule{0pt}{10pt}
%			\Checkmark & \Checkmark & \Checkmark & \Checkmark & 61.6 \\ \rule{0pt}{10pt}
%			\Checkmark & \Checkmark & \Checkmark & \XSolidBrush & 61.3 \\ \rule{0pt}{10pt}
%			\Checkmark & \Checkmark & \XSolidBrush & \XSolidBrush & 60.5 \\ \rule{0pt}{10pt}
%			\Checkmark & \XSolidBrush & \XSolidBrush & \XSolidBrush & 60.4 \\ \bottomrule[1.0pt] \rule{0pt}{10pt}			
%		\end{tabular}
%	}
%\end{table}

\begin{table}[t]
	\caption{Ablation Studies on the Global Encoder in Different Encoding Layers}
	\label{table_example}
	\resizebox{1.0\linewidth}{!}{
		\begin{tabular}{l|c|c|c|c}
			\toprule[1.0pt]
			Encoding layer & 1st $\sim$ 4th & 1st $\sim$ 3rd & 1st $\sim$ 2nd & 1st only \\ \midrule[0.3pt]
			mIoU (\%) & 61.6 & 61.3 & 60.5 & 60.4 \\ \bottomrule[1.0pt]
		\end{tabular}
	}
\end{table}

\subsection{Complexity Analysis}

The complexity of the network is crucial for practical applications. As shown in Table \textcolor[rgb]{1,0,0}{\uppercase\expandafter{\romannumeral5}}, we conducted a complexity analysis on the SemanticKITTI dataset. We compare the time and memory costs of our model against other models, including model parameter, FLOPs (Floating Point Operations), GPU memory usage, and the elapsed to process $10^5$ points in a point cloud. From the table, it can be seen that our method demonstrates advantages in terms of mIoU, FLOPs, and GPU memory. PointNet\cite{qi2017pointnet} boasts the fastest processing speed, yet its encoding layers lack 
\begin{table}[t]
	\caption{Complexity Analysis of Different Methods on SemanticKITTI}
	\label{table_example}
	\resizebox{1.0\linewidth}{!}{
		\begin{tabular}{cccccc}
			\toprule[1.5pt]
			Method & Param (M) & FLOPs (G) & GPU Mem. (GB) & \makecell{Time cost \\ (sec. / $10^5$ points)} & mIoU (\%) \\ \midrule[0.5pt]
			PointNet\cite{qi2017pointnet} & 3.53 & 104.87 & 6.02 & \textbf{0.0239} & 14.6\\ \rule{0pt}{10pt}
			PointNet++\cite{qi2017pointnet++} & \textbf{0.95} & 16.22 & 20.71 & 14.1145 & 20.1\\ \rule{0pt}{10pt}
			PointASNL\cite{yan2020pointasnl} & 3.98 & 470.85 & 28.88 & 12.2996 & 46.8\\ \rule{0pt}{10pt}
			RandLA-Net\cite{hu2021learning} & 1.24 & 14.55 & 5.47 & 0.0357 & 55.9\\ \rule{0pt}{10pt}
			SWCF-Net (Ours) & 3.36 & \textbf{13.39} & \textbf{5.17} & 0.0435 & \textbf{60.5}\\ \bottomrule[1.5pt]
			
		\end{tabular}
	}
\end{table}
a downsampling operation on point clouds, which in turn results in a substantial computational overhead. PointNet++\cite{qi2017pointnet++} has the smallest parameter count, but it utilizes Farthest Point Sampling between layers when processing large-scale point clouds, which consequently leads to excessive processing time. PointASNL\cite{yan2020pointasnl} adopts a combined strategy of local and global feature extraction for point clouds. However, its global feature extractor utilizes a traditional attention mechanism, which in turn results in substantial GPU memory consumption and computational requirements. To mitigate the excessively high computational demand of the attention mechanism when processing large-scale point clouds, our method proposes the Average Transformer. This innovation enables us to significantly reduce both GPU memory usage and computational requirements. Moreover, compared to RandLA-Net which also employs random downsampling, our method achieves better segmentation results due to its consideration of the underlying semantic contextual information in large-scale point clouds. Our method is capable of processing around $10^5$ points in approximately 0.04 seconds, thus attaining a more favorable trade-off between accuracy and speed.

\section{CONCLUSIONS}

In this paper, we design a point-based network named SWCF-Net for efficient large-scale point cloud semantic segmentation. To fully utilize the underlying geometric feature and global semantic context in large-scale point clouds, we propose SWConv for extracting local geometric features, and introduce the lightweight attention module entitled Average Transformer to capture global semantic context and long-range dependencies. Finally, to effectively integrate both types of features, we propose an Orthogonal Fusion module that performs redundancy-reducing fusion of local and global features. Our SWCF-Net improves the performance of semantic segmentation by jointly considering local and global features. Experimental results indicate that our proposed SWCF-Net effectively and accurately performs semantic segmentation on large-scale point clouds.

\balance

\bibliographystyle{ieeetran}
\bibliography{pc-ref.bib}

\begin{thebibliography}{10}
\providecommand{\url}[1]{#1}
\csname url@samestyle\endcsname
\providecommand{\newblock}{\relax}
\providecommand{\bibinfo}[2]{#2}
\providecommand{\BIBentrySTDinterwordspacing}{\spaceskip=0pt\relax}
\providecommand{\BIBentryALTinterwordstretchfactor}{4}
\providecommand{\BIBentryALTinterwordspacing}{\spaceskip=\fontdimen2\font plus
\BIBentryALTinterwordstretchfactor\fontdimen3\font minus
  \fontdimen4\font\relax}
\providecommand{\BIBforeignlanguage}[2]{{%
\expandafter\ifx\csname l@#1\endcsname\relax
\typeout{** WARNING: IEEEtran.bst: No hyphenation pattern has been}%
\typeout{** loaded for the language `#1'. Using the pattern for}%
\typeout{** the default language instead.}%
\else
\language=\csname l@#1\endcsname
\fi
#2}}
\providecommand{\BIBdecl}{\relax}
\BIBdecl

\bibitem{guo2020deep}
Y.~Guo, H.~Wang, Q.~Hu, H.~Liu, L.~Liu, and M.~Bennamoun, ``Deep learning for
  3d point clouds: A survey,'' \emph{IEEE transactions on pattern analysis and
  machine intelligence}, vol.~43, no.~12, pp. 4338--4364, 2020.

\bibitem{qi2017pointnet++}
C.~R. Qi, L.~Yi, H.~Su, and L.~J. Guibas, ``Pointnet++: Deep hierarchical
  feature learning on point sets in a metric space,'' \emph{Advances in neural
  information processing systems}, vol.~30, 2017.

\bibitem{ma2022rethinking}
X.~Ma, C.~Qin, H.~You, H.~Ran, and Y.~Fu, ``Rethinking network design and local
  geometry in point cloud: A simple residual mlp framework,'' \emph{arXiv
  preprint arXiv:2202.07123}, 2022.

\bibitem{landrieu2018large}
L.~Landrieu and M.~Simonovsky, ``Large-scale point cloud semantic segmentation
  with superpoint graphs,'' in \emph{Proceedings of the IEEE conference on
  computer vision and pattern recognition}, 2018, pp. 4558--4567.

\bibitem{hu2021learning}
Q.~Hu, B.~Yang, L.~Xie, S.~Rosa, Y.~Guo, Z.~Wang, N.~Trigoni, and A.~Markham,
  ``Learning semantic segmentation of large-scale point clouds with random
  sampling,'' \emph{IEEE Transactions on Pattern Analysis and Machine
  Intelligence}, vol.~44, no.~11, pp. 8338--8354, 2021.

\bibitem{qiu2021semantic}
S.~Qiu, S.~Anwar, and N.~Barnes, ``Semantic segmentation for real point cloud
  scenes via bilateral augmentation and adaptive fusion,'' in \emph{Proceedings
  of the IEEE/CVF Conference on Computer Vision and Pattern Recognition}, 2021,
  pp. 1757--1767.

\bibitem{ruan2023combining}
J.~Ruan, L.~He, Y.~Guan, and H.~Zhang, ``Combining scene coordinate regression
  and absolute pose regression for visual relocalization,'' in \emph{2023 IEEE
  International Conference on Robotics and Automation (ICRA)}.\hskip 1em plus
  0.5em minus 0.4em\relax IEEE, 2023, pp. 11\,749--11\,755.

\bibitem{yan2020pointasnl}
X.~Yan, C.~Zheng, Z.~Li, S.~Wang, and S.~Cui, ``Pointasnl: Robust point clouds
  processing using nonlocal neural networks with adaptive sampling,'' in
  \emph{Proceedings of the IEEE/CVF conference on computer vision and pattern
  recognition}, 2020, pp. 5589--5598.

\bibitem{qi2017pointnet}
C.~R. Qi, H.~Su, K.~Mo, and L.~J. Guibas, ``Pointnet: Deep learning on point
  sets for 3d classification and segmentation,'' in \emph{Proceedings of the
  IEEE conference on computer vision and pattern recognition}, 2017, pp.
  652--660.

\bibitem{wang2019dynamic}
Y.~Wang, Y.~Sun, Z.~Liu, S.~E. Sarma, M.~M. Bronstein, and J.~M. Solomon,
  ``Dynamic graph cnn for learning on point clouds,'' \emph{ACM Transactions on
  Graphics (tog)}, vol.~38, no.~5, pp. 1--12, 2019.

\bibitem{thomas2019kpconv}
H.~Thomas, C.~R. Qi, J.-E. Deschaud, B.~Marcotegui, F.~Goulette, and L.~J.
  Guibas, ``Kpconv: Flexible and deformable convolution for point clouds,'' in
  \emph{Proceedings of the IEEE/CVF international conference on computer
  vision}, 2019, pp. 6411--6420.

\bibitem{zhao2021point}
H.~Zhao, L.~Jiang, J.~Jia, P.~H. Torr, and V.~Koltun, ``Point transformer,'' in
  \emph{Proceedings of the IEEE/CVF international conference on computer
  vision}, 2021, pp. 16\,259--16\,268.

\bibitem{tatarchenko2018tangent}
M.~Tatarchenko, J.~Park, V.~Koltun, and Q.-Y. Zhou, ``Tangent convolutions for
  dense prediction in 3d,'' in \emph{Proceedings of the IEEE conference on
  computer vision and pattern recognition}, 2018, pp. 3887--3896.

\bibitem{yin2023dcnet}
F.~Yin, Z.~Huang, T.~Chen, G.~Luo, G.~Yu, and B.~Fu, ``Dcnet: Large-scale point
  cloud semantic segmentation with discriminative and efficient feature
  aggregation,'' \emph{IEEE Transactions on Circuits and Systems for Video
  Technology}, 2023.

\bibitem{du2022novel}
Z.~Du, H.~Ye, and F.~Cao, ``A novel local-global graph convolutional method for
  point cloud semantic segmentation,'' \emph{IEEE Transactions on Neural
  Networks and Learning Systems}, 2022.

\bibitem{zhao2023large}
Y.~Zhao, X.~Ma, B.~Hu, Q.~Zhang, M.~Ye, and G.~Zhou, ``A large-scale point
  cloud semantic segmentation network via local dual features and global
  correlations,'' \emph{Computers \& Graphics}, vol. 111, pp. 133--144, 2023.

\bibitem{ren2015faster}
S.~Ren, K.~He, R.~Girshick, and J.~Sun, ``Faster r-cnn: Towards real-time
  object detection with region proposal networks,'' \emph{Advances in neural
  information processing systems}, vol.~28, 2015.

\bibitem{strudel2021segmenter}
R.~Strudel, R.~Garcia, I.~Laptev, and C.~Schmid, ``Segmenter: Transformer for
  semantic segmentation,'' in \emph{Proceedings of the IEEE/CVF international
  conference on computer vision}, 2021, pp. 7262--7272.

\bibitem{wu2019pointconv}
W.~Wu, Z.~Qi, and L.~Fuxin, ``Pointconv: Deep convolutional networks on 3d
  point clouds,'' in \emph{Proceedings of the IEEE/CVF Conference on computer
  vision and pattern recognition}, 2019, pp. 9621--9630.

\bibitem{bartlett2002rademacher}
P.~L. Bartlett and S.~Mendelson, ``Rademacher and gaussian complexities: Risk
  bounds and structural results,'' \emph{Journal of Machine Learning Research},
  vol.~3, no. Nov, pp. 463--482, 2002.

\bibitem{li2016filter}
X.~Li, F.~Li, X.~Fern, and R.~Raich, ``Filter shaping for convolutional neural
  networks,'' in \emph{International Conference on Learning Representations},
  2016.

\bibitem{vaswani2017attention}
A.~Vaswani, N.~Shazeer, N.~Parmar, J.~Uszkoreit, L.~Jones, A.~N. Gomez,
  {\L}.~Kaiser, and I.~Polosukhin, ``Attention is all you need,''
  \emph{Advances in neural information processing systems}, vol.~30, 2017.

\bibitem{sharma2023truth}
P.~Sharma, J.~T. Ash, and D.~Misra, ``The truth is in there: Improving
  reasoning in language models with layer-selective rank reduction,''
  \emph{arXiv preprint arXiv:2312.13558}, 2023.

\bibitem{han2023flatten}
D.~Han, X.~Pan, Y.~Han, S.~Song, and G.~Huang, ``Flatten transformer: Vision
  transformer using focused linear attention,'' in \emph{Proceedings of the
  IEEE/CVF International Conference on Computer Vision}, 2023, pp. 5961--5971.

\bibitem{wang2022pvt}
W.~Wang, E.~Xie, X.~Li, D.-P. Fan, K.~Song, D.~Liang, T.~Lu, P.~Luo, and
  L.~Shao, ``Pvt v2: Improved baselines with pyramid vision transformer,''
  \emph{Computational Visual Media}, vol.~8, no.~3, pp. 415--424, 2022.

\bibitem{reinke2022locus}
A.~Reinke, M.~Palieri, B.~Morrell, Y.~Chang, K.~Ebadi, L.~Carlone, and A.-A.
  Agha-Mohammadi, ``Locus 2.0: Robust and computationally efficient lidar
  odometry for real-time 3d mapping,'' \emph{IEEE Robotics and Automation
  Letters}, vol.~7, no.~4, pp. 9043--9050, 2022.

\bibitem{zhu2022curvature}
L.~Zhu, W.~Chen, X.~Lin, L.~He, and Y.~Guan, ``Curvature-variation-inspired
  sampling for point cloud classification and segmentation,'' \emph{IEEE Signal
  Processing Letters}, vol.~29, pp. 1868--1872, 2022.

\bibitem{zhang2022patchformer}
C.~Zhang, H.~Wan, X.~Shen, and Z.~Wu, ``Patchformer: An efficient point
  transformer with patch attention,'' in \emph{Proceedings of the IEEE/CVF
  Conference on Computer Vision and Pattern Recognition}, 2022, pp.
  11\,799--11\,808.

\bibitem{milioto2019rangenet++}
A.~Milioto, I.~Vizzo, J.~Behley, and C.~Stachniss, ``Rangenet++: Fast and
  accurate lidar semantic segmentation,'' in \emph{2019 IEEE/RSJ international
  conference on intelligent robots and systems (IROS)}.\hskip 1em plus 0.5em
  minus 0.4em\relax IEEE, 2019, pp. 4213--4220.

\bibitem{xu2020squeezesegv3}
C.~Xu, B.~Wu, Z.~Wang, W.~Zhan, P.~Vajda, K.~Keutzer, and M.~Tomizuka,
  ``Squeezesegv3: Spatially-adaptive convolution for efficient point-cloud
  segmentation,'' in \emph{Computer Vision--ECCV 2020: 16th European
  Conference, Glasgow, UK, August 23--28, 2020, Proceedings, Part XXVIII
  16}.\hskip 1em plus 0.5em minus 0.4em\relax Springer, 2020, pp. 1--19.

\bibitem{behley2019semantickitti}
J.~Behley, M.~Garbade, A.~Milioto, J.~Quenzel, S.~Behnke, C.~Stachniss, and
  J.~Gall, ``Semantickitti: A dataset for semantic scene understanding of lidar
  sequences,'' in \emph{Proceedings of the IEEE/CVF international conference on
  computer vision}, 2019, pp. 9297--9307.

\bibitem{tan2020toronto}
W.~Tan, N.~Qin, L.~Ma, Y.~Li, J.~Du, G.~Cai, K.~Yang, and J.~Li, ``Toronto-3d:
  A large-scale mobile lidar dataset for semantic segmentation of urban
  roadways,'' in \emph{Proceedings of the IEEE/CVF conference on computer
  vision and pattern recognition workshops}, 2020, pp. 202--203.

\bibitem{ma2019multi}
L.~Ma, Y.~Li, J.~Li, W.~Tan, Y.~Yu, and M.~A. Chapman, ``Multi-scale point-wise
  convolutional neural networks for 3d object segmentation from lidar point
  clouds in large-scale environments,'' \emph{IEEE Transactions on Intelligent
  Transportation Systems}, vol.~22, no.~2, pp. 821--836, 2019.

\bibitem{chen2023feature}
J.~Chen, Y.~Chen, and C.~Wang, ``Feature graph convolution network with
  attentive fusion for large-scale point clouds semantic segmentation,''
  \emph{IEEE Geoscience and Remote Sensing Letters}, 2023.

\end{thebibliography}

\end{document}